\documentclass[runningheads]{llncs}
\usepackage{graphicx}
\usepackage{multirow}
\usepackage{makecell}
\usepackage{diagbox}
\usepackage{footnote}
\usepackage{amssymb}
\usepackage{amsmath}
\usepackage{color}
\usepackage{ulem}
\usepackage{xspace}
\newcommand{\eg}{{\it e.g.,}\xspace}
\newcommand{\ie}{{\it i.e.,}\xspace}
\usepackage[flushleft]{threeparttable}
\usepackage[misc,geometry]{ifsym}

\begin{document}
%

\title{VSR: A Unified Framework for Document Layout Analysis combining Vision, Semantics and Relations}

\titlerunning{VSR}
%

\author{Peng Zhang\inst{1}
	\and Can Li\inst{1}
	\and Liang Qiao\inst{1}
	\and Zhanzhan Cheng\inst{2,1}
	\and Shiliang Pu\inst{1}
	\and Yi Niu\inst{1}
	\and Fei Wu\inst{2}}
\authorrunning{P. Zhang, C. Li, Z. Cheng et al.}
%
\institute{Hikvision Research Institute, China \\
	\email{\{zhangpeng23, lican9, qiaoliang6, chengzhanzhan, pushiliang.hri, niuyi\}@hikvision.com}
	\and
	Zhejiang University, China \\
	\email{wufei@cs.zju.edu.cn}}

\maketitle
%
\begin{abstract}
	Document layout analysis is crucial for understanding document structures.
	On this task, \textit{vision} and \textit{semantics} of documents, and \textit{relations} between layout components contribute to the understanding process.
	Though many works have been proposed to exploit the above information, they show unsatisfactory results.
	NLP-based methods model layout analysis as a sequence labeling task and show insufficient capabilities in layout modeling.
	CV-based methods model layout analysis as a detection or segmentation task, but bear limitations of inefficient modality fusion and lack of relation modeling between layout components.
	To address the above limitations, we propose a unified framework VSR for document layout analysis, combining \textit{vision}, \textit{semantics} and \textit{relations}.
	VSR supports both NLP-based and CV-based methods.
	Specifically, we first introduce \textit{vision} through document image and \textit{semantics} through text embedding maps.
	Then, modality-specific visual and semantic features are extracted using a \textit{two-stream} network, which are \textit{adaptively} fused to make full use of complementary information.
	Finally, given component candidates, a \textit{relation module} based on graph neural network is incorported to model relations between components and output final results.
	On three popular benchmarks, VSR outperforms previous models by large margins.
	Code will be released soon.
	
	\keywords{Vision \and Semantics \and Relations \and Document layout analysis.}
\end{abstract}
\section{Introduction}
\label{introduction}
Document layout analysis is a crucial step in automatic document understanding and enables many important applications, such as document retrieval~\cite{DBLP:journals/csur/BinMakhashenM20}, digitization~\cite{DBLP:conf/icpr/CorbelliBGC16} and editing.
Its goal is to identify the regions of interest in unstructured document and recognize the role of each region.
This task is challenging due to the diversity and complexity of document layouts. 

Many deep learning models have been proposed on this task in both computer vision (CV) and natural language processing (NLP) communities.
Most of them consider either only visual features~\cite{DBLP:conf/icdar/HeCPKG17,DBLP:conf/icdar/ChenSLHI15,DBLP:conf/das/WickP18,DBLP:conf/icfhr/GatosLS14,DBLP:conf/icip/VoL16,DBLP:conf/icfhr/ZagorisPG14,DBLP:conf/cvpr/LiWTZBMMSF20,DBLP:conf/icdar/LiYXLOL19} or only semantic features~\cite{DBLP:conf/icdar/Conway93,DBLP:journals/pami/KrishnamoorthyNSV93,DBLP:conf/iccv/ShilmanLV05}. 
However, information from both modalities could help recognize the document layout better. Some regions (\eg {Figure}, {Table}) can be easily identified by visual features, while semantic features are important for separating visually similar regions (\eg {Abstract} and {Paragraph}).
Therefore, some recent efforts try to combine both modalities~\cite{DBLP:conf/wacv/AggarwalSGK20,DBLP:conf/coling/LiXCHWLZ20,DBLP:conf/cvpr/YangYAKKG17,DBLP:journals/corr/abs-2002-06144}.  Here we summarize them into two categories.

\begin{figure}[t]
	\centering
	\includegraphics[width=\textwidth]{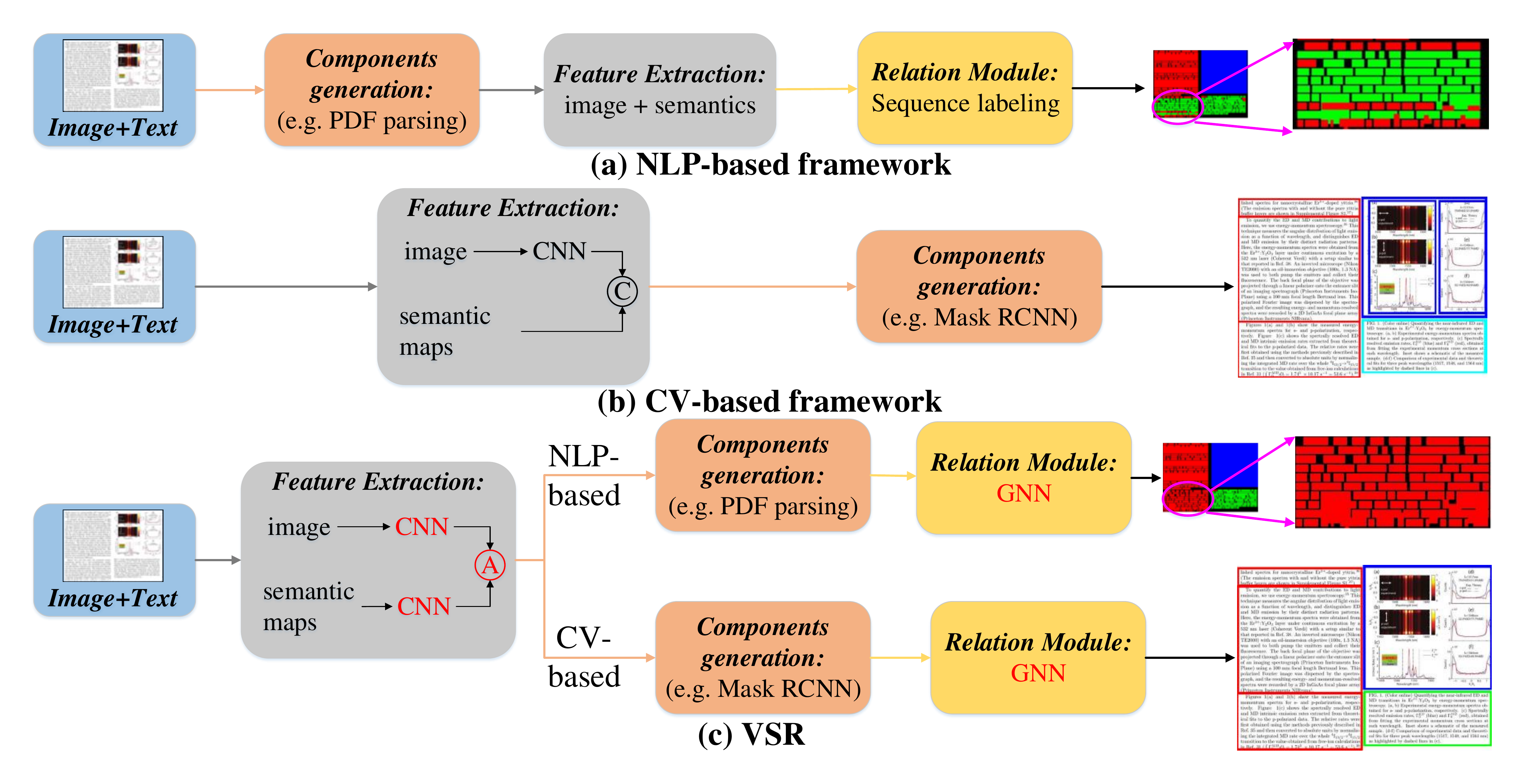}
	\caption{Comparison of multimodal document layout analysis frameworks.
		VSR supports both NLP-based and CV-based frameworks.
		$\copyright$ and {\color{red}\textcircled{A}}
		denote \textit{concatenation} and \textit{adaptive aggregation} fusion strategies.
		Different colored regions in the prediction results indicate different semantic labels ({\color{red}{Paragraph}, \color{blue}{Figure}, \color{green}{Figure Caption}, \color{cyan}{Table Caption}}).}
	\label{figure:architectore_comp}
\end{figure}

{NLP-based methods} (Fig~\ref{figure:architectore_comp} (a)) {model layout analysis as a sequence labeling task and }apply a bottom-up strategy.
They first serialize texts into 1D token sequence\footnote[1]{In the rest of this paper, we assume text is available. There are tools available to extract text from PDF documents (\eg PDFMiner~\cite{shinyama2015pdfminer}) and document images (\eg OCR engine~\cite{DBLP:conf/icdar/Smith07}).}.
Then, using both semantic and visual features (such as coordinates and image embedding) of each token, {they} determine token labels sequentially {through a sequence labeling model}.
{However, NLP-based methods show insufficient capabilities in layout modeling. For example in Fig.~\ref{figure:architectore_comp}(a), all texts in a paragraph should have consistent semantic labels ({{\color{red}Paragraph}}), but some of them are recognized as {{\color{green}Figure Caption}}, which are the labels of adjacent texts.}

{CV-based methods} (Fig~\ref{figure:architectore_comp} (b)) {model layout analysis as object detection or segmentation task, and} apply a top-down strategy.
{They first extract visual features by convolutional neural network and introduce semantic features through text embedding maps (at sentence-level~\cite{DBLP:conf/cvpr/YangYAKKG17} \textit{or} character-level~\cite{DBLP:journals/corr/abs-2002-06144}), which are directly concatenated as the representation of document. Then, detection or segmentation models (\eg Mask RCNN~\cite{DBLP:conf/iccv/HeGDG17}) are used to generate layout component candidates (coordinates and semantic labels).}
While capturing spatial information {better} compared to {NLP-based methods}, {CV-based methods} still have 3 limitations: 
(1) \textit{limited semantics.}
Semantic information are embedded in {text} at different granularities, including characters (or words) and sentences, which could help identify different document elements. {For example, character-level features are better for recognizing components which need less context (\eg Author) while sentence-level features are better for contextual components (\eg Table caption).
Exploiting semantics at one granularity could not achieve optimal performances.}
(2) {\textit{simple and heuristic modality fusion strategy}. Features from different modalities contribute differently to component recognition. Visual features contribute more to recognizing visually rich components (such as {Figure} and {Table}), while semantic features are better at distinguishing text-based components ({Abstract} and {Paragraph}). 
Simple and heuristic modality fusion by concatenation can not fully make use of complementary information between two modalities.}
(3) \textit{lack of relation modeling between components}. Strong relations exist in documents. For example, ``{Figure}'' and ``{Figure Caption}'' often appear together, {and ``{Paragraph}''s have aligned bounding box coordinates.} Such relations could be utilized to {boost layout analysis performances.}

In this paper, we propose a unified framework VSR for document layout analysis, combining \textit{\textbf{V}ision}, \textit{\textbf{S}emantics} and \textit{\textbf{R}elation modeling}, as shown in Fig~\ref{figure:architectore_comp} (c).
{This framework can be applied to both NLP-based and CV-based methods.}
{First, documents are fed into VSR in the form of images (vision) and text embedding maps (semantics at both character-level and sentence-level).}
{Then, modality-specific visual and semantic features are extracted through a two-stream network, which are {effectively} combined later in a multi-scale adaptive aggregation module.
Finally, a GNN(Graph Neural Network)-based relation module is incorporated to model relations between component candidates, and generate final results.
Specifically, for NLP-based methods, text tokens serve as the component candidates and relation module predicts their semantic labels. 
While for CV-based methods, component candidates are proposed by detection or segmentation model (\eg Faster RCNN/ Mask RCNN) and relation module generates their refined coordinates and semantic labels.}

Our work makes four key contributions:
\begin{itemize}
	\item We propose a unified framework VSR for document layout analysis, combining \textit{vision}, \textit{semantics} and \textit{relations} in documents.
    \item To exploit vision and semantics effectively, we propose a \textit{two-stream} network to extract modality-specific visual and semantic features, and fuse them \textit{adaptively} through an adaptive aggregation module. Besides, we also explore document semantics at different granularities.
	\item A GNN-based relation module is incorporated to model relations between document components, and it supports relation modeling in both NLP-based and CV-based methods.
	\item We perform extensive evaluations of VSR, and on three public benchmarks, VSR shows significant improvements compared to previous models.
\end{itemize}
\section{Related Works}
\subsubsection{Document Layout Analysis.}
{
	In this paper, we try to review layout analysis works from the perspective of {modality} used, namely, \textit{unimodal layout analysis} and \textit{multimodal layout analysis}.
	
	\textit{Unimodal layout analysis} exploits either only visual features~\cite{DBLP:conf/icdar/LiYXLOL19,DBLP:conf/cvpr/LiWTZBMMSF20} (document image) or only semantic features (document texts) to understand document structures.
	Using visual features, several works~\cite{DBLP:conf/icdar/ChenSLHI15,DBLP:conf/das/WickP18} have been proposed to apply CNN to segment various objects, \eg text blocks~\cite{DBLP:conf/icfhr/GatosLS14}, text lines~\cite{DBLP:conf/icip/VoL16,DBLP:conf/icdar/LeeHOU19}, words~\cite{DBLP:conf/icfhr/ZagorisPG14}, figures or tables~\cite{DBLP:conf/icdar/HeCPKG17,DBLP:conf/jcdl/SiegelLPA18}.
	At the same time, there are also methods~\cite{DBLP:conf/icdar/Conway93,DBLP:journals/pami/KrishnamoorthyNSV93,DBLP:conf/iccv/ShilmanLV05} which try to address the layout analysis problem using semantic features.
	However, all the above methods are strictly restricted to visual \textit{or} semantic features, and thus are not able to exploit complementary information from other modalities.
	
	\textit{Multimodal layout analysis} tries to combine information from both visual and semantic modalities.
	Related methods can be further divided into two categories, NLP-based and CV-based methods.
	NLP-based methods work on low-level elements (\eg tokens) and model layout analysis as a sequence labeling task.
	MMPAN~\cite{DBLP:conf/wacv/AggarwalSGK20} is presented to recognize form structures.
	DocBank~\cite{DBLP:conf/coling/LiXCHWLZ20} is proposed as a large scale dataset of multimodal layout analysis and several NLP baselines have been released.
	However, the above methods show insufficient capabilities in layout modeling.
	CV-based methods introduce document semantics through text embedding maps, and model layout analysis as object detection or segmentation task.
	MFCN~\cite{DBLP:conf/cvpr/YangYAKKG17} introduces sentence granularity semantics and inserts the text embedding maps at the decision-level (end of network), while \textit{dhSegment$^T$}\footnote[2]{\textit{dhSegment$^T$} means \textit{dhSegment} with inputs of image and text embedding maps.}~\cite{DBLP:journals/corr/abs-2002-06144} introduces character granularity semantics and inserts text embedding maps at the input-level.
	Though showing great success, the above methods also bear the following limitations: limited semantics used, simple modality fusion strategy and lack of relation modeling between components.
	
	To remedy the above limitations, we propose a unified framework VSR to exploit vision, semantics and relations in documents.
}

\subsubsection{Two-stream networks.}
 {Two-stream networks are widely used to combine features in different modalities or representations~\cite{DBLP:journals/pami/BaltrusaitisAM19} effectively.}
In action recognition, two-stream networks are used to capture the complementary \textit{spatial} and \textit{temporal} information~\cite{DBLP:conf/cvpr/FeichtenhoferPZ16}.
In RGB-D saliency detection, the complete representations are fused from the deep features of the \textit{RGB} stream and \textit{depth} stream~\cite{DBLP:journals/tcyb/HanCLYL18}.
Also, two-stream networks are used to fuse different features of same input sample in sound event classification and image recognition~\cite{DBLP:conf/iccv/LinRM15}.
Motivated by their successes, we apply two-stream networks to capture complementary \textit{vision} and \textit{semantics} information in documents.

\subsubsection{Relation modeling.}
Relation modeling is a broad topic and has been studing for decades.
In natural language processing, dependencies between sequential texts are captured through \textit{RNN}~\cite{DBLP:journals/neco/HochreiterS97} or \textit{Transformer}~\cite{DBLP:conf//VaswaniSPUJGKP17} architectures.
In computer vision,
{non-local networks~\cite{DBLP:conf/cvpr/0004GGH18} and relation networks~\cite{DBLP:conf/cvpr/HuGZDW18} are presented to model long-range dependencies between pixels and objects.}
Besides, in document image processing, {relations between text and layout~\cite{DBLP:conf/kdd/XuL0HW020}  or relations between document entities~\cite{DBLP:conf/icpr/YuLQG020,DBLP:conf/naacl/LiuGZZ19,DBLP:conf/mm/ZhangXCP0QNW20}  are explored.}
As to multimodal layout analysis, NLP-based methods model it as a sequence labeling task and use \textit{RNN} to capture component relations, while CV-based methods model it as object detection task but lack relation modeling between layout components.
In this paper, we propose a GNN-based relation module, supporting relation modeling in both NLP-based or CV-based methods.

\section{Methodology}
\label{methodology}
\begin{figure}[t]
	\centering
	\includegraphics[width=\textwidth]{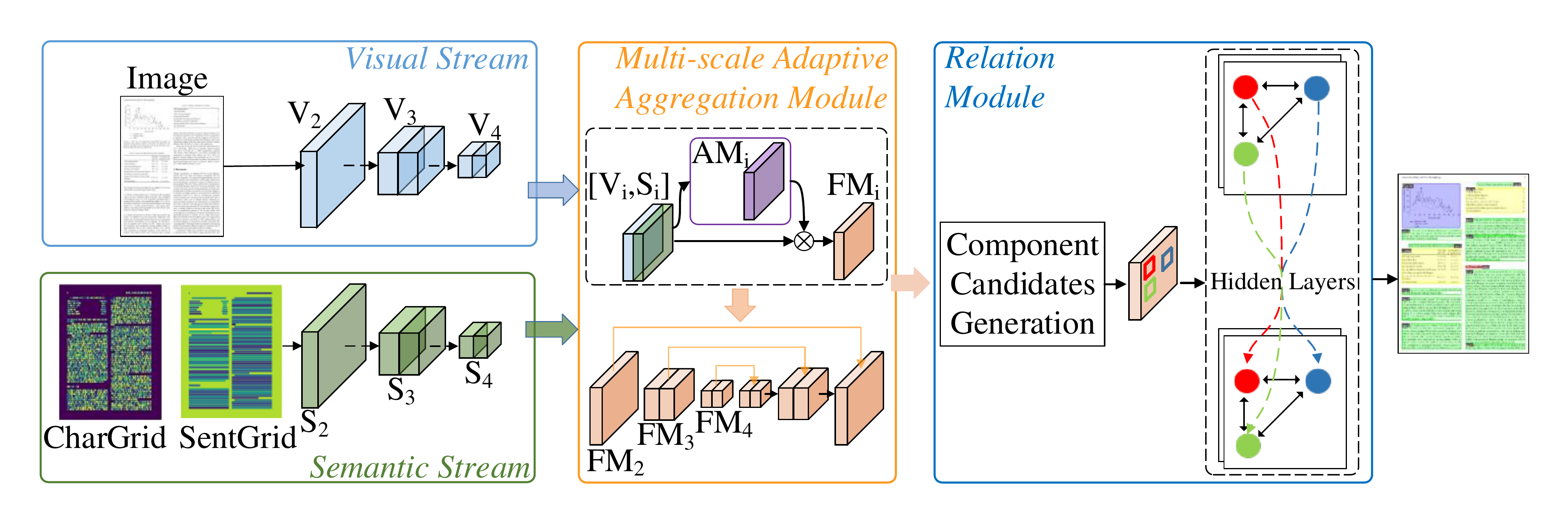}
	\caption{The architecture overview of VSR. Best viewed in color.}
	\label{figure:system_architecture}
\end{figure}

\subsection{Architecture Overview}
\label{overall_architecture}
{Our proposed framework has three parts:}
two-stream ConvNets, a multi-scale adaptive aggregation module and a relation module{ (as shown in Fig~\ref{figure:system_architecture})}.
{First, a two-stream convolutional network extracts modality-specific visual and semantic features, where 
{visual stream and semantic stream take images and text embedding maps as input, respectively (Sec \ref{two_stream_convnets}).}
Next, instead of simply concatenating the visual and semantic features, we aggregate them via a multi-scale adaptive aggregation module (Sec \ref{sec_maa}).
{Then, a set of component candidates are generated.}
Finally, a relation module is incorporated to model relations {among those candidates} and generate final results (Sec \ref{sec_rm}).

{Notice that multimodal layout analysis can be modeled as sequence labeling (NLP-based methods) or object detection tasks (CV-based methods). 
Our framework supports both modeling types. The only difference is what the component candidates are and how to generate them.
Component candidates are low-level elements (\eg text tokens) in NLP-based methods and can be generated by parsing PDFs, while candidates
are high-level elements (regions) generated by detection or segmentation model (\eg Mask RCNN) in CV-based methods. In the rest of this paper, we will illustrate how VSR is applied to CV-based methods, 
and show it can be easily adapted to NLP-based methods in experiments on DocBank benchmark (Sec~\ref{main_result}).
}}

\subsection{Two-stream ConvNets}
\label{two_stream_convnets}
{{CNN is known to be good at learning deep features. However, previous multimodal layout analysis works~\cite{DBLP:conf/cvpr/YangYAKKG17,DBLP:journals/corr/abs-2002-06144} only apply it to extract visual features. Text embedding maps are directly used as semantic features.}
This single-stream network design could not make full use of document semantics.
Motivated by great success of two-stream network in various multimodal applications \cite{DBLP:conf/cvpr/FeichtenhoferPZ16,DBLP:conf/iccv/LinRM15}, we apply it to extract deep visual and semantic features.

\subsubsection{Visual stream ConvNet.}
This stream directly takes document images as input and extracts multi-scale deep features using CNN backbones like ResNet \cite{DBLP:conf/cvpr/HeZRS16}.
Specifically, for an input image $x \in \mathbb{R}^{H \times W \times 3}$, multi-scale features maps (denoted by $\left\{V_2, V_3, V_4, V_5\right\}$) are extracted, 
where each $V_i \in \mathbb{R}^{\frac{H}{2^i} \times \frac{W}{2^i} \times C_i^V}$.
$H$ and $W$ are the height and width of input image $x$, $C_i^V$ is the channel dimension of feature map $V_i$, and $V_0=x$.

\subsubsection{Semantic stream ConvNet.}
Similar to~\cite{DBLP:conf/cvpr/YangYAKKG17,DBLP:journals/corr/abs-2002-06144}, we introduce document semantics through text embedding maps $S_0 \in \mathbb{R}^{H \times W \times C_0^S}$, which are the input of semantic stream ConvNet.
$S_0$ have same spatial sizes with document image $x$ ($V_0$) and $C_0^S$ denotes the initial channel dimension.
This type of representation not only encodes text {content}, but also preserves the 2D layout of a document.
{Previously, only semantics at one granularity is used (character-level~\cite{DBLP:journals/corr/abs-2002-06144} \textit{or} sentence-level\footnote[3]{Sentence is a group of words or phrases, which usually ends with a period, question mark or exclamation point. For simplicity, we approximate it with text lines.}~\cite{DBLP:conf/cvpr/YangYAKKG17}). 
{However, semantics at different granularities contribute to identification of different components.}
Thus, $S_0$ consists of both character and sentence level semantics.
Next, we show how we build text embedding maps $S_0$.
}

The characters and sentences of a document page are denoted as $\mathbb{D}_c=\left\{\left( c_k, b_k^c\right) | k=0,\cdots, n\right\}$ and $\mathbb{D}_s=\left\{\left( s_k, b_k^s\right) | k=0,\cdots, m\right\}$, where $n$ and $m$ are the total number of characters and sentences.
$c_k$ and $b_k^c=\left(x_0, y_0, x_1, y_1\right)$ are the $k$-th character and its associated box, where $(x_0, y_0)$ and $(x_1, y_1)$ are top-left and bottom-right pixel coordinates.
{Similarly,  $s_k$ and $b_k^s$ are the $k$-th sentence and its box location}.
Next, character embedding maps $CharGrid \in \mathbb{R}^{H \times W \times C_0^S}$ and sentence embedding maps $SentGrid \in \mathbb{R}^{H \times W \times C_0^S}$ can be constructed as follows.
\begin{equation}
CharGrid_{ij} = \left\{
\begin{array}{ll}
E^c(c_k) & \mbox{if}\ (i,j) \in b_k^c \\
0 & \mbox{otherwise}
\end{array}
\right.
\end{equation}

\begin{equation}
SentGrid_{ij} = \left\{
\begin{array}{ll}
E^s(s_k) & \mbox{if}\ (i,j) \in b_k^s \\
0 & \mbox{otherwise}
\end{array}
\right.
\end{equation}

All pixels in each $b_k^c$ ($b_k^s$) share the same character (sentence) embedding vector.
$E^c$ and $E^s$ are the mapping functions of $c_k\rightarrow\mathbb{R}^{C^S_0}$ and $s_k\rightarrow\mathbb{R}^{C^S_0}$.
In our implementation, $E^c$ is a typical word embedding layer and  {we adopt pretrained language model BERT~\cite{DBLP:conf/naacl/DevlinCLT19} as $E^s$.}
Finally, the {text} embedding maps $S_0$ can be {constructed by applying LayerNorm normalization to the summation of  $CharGrid$ and $SentGrid$, as shown in Eq.(\ref{eq-s0})}.
\begin{equation}
S_0 = LayerNorm\left(CharGrid + SentGrid\right)
\label{eq-s0}
\end{equation}

Similar to the visual stream, semantic stream ConvNet then takes text embedding maps $S_0$ as input and extracts multi-scale features $\left\{S_2, S_3, S_4, S_5\right\}$, which have the same spatial sizes and channel dimension with $\left\{V_2, V_3, V_4, V_5\right\}$.

\subsection{Multi-scale Adaptive Aggregation}
\label{sec_maa}
{Features from different modalities are important for identifying different objects. 
Modality fusion strategy should adaptively aggregate visual and semantic features. 
{Thus,} we design a multi-scale adaptive aggregation module that learns an attention map to combine visual features $\left\{V_2, V_3, V_4, V_5\right\}$ and semantic features $\left\{S_2, S_3, S_4, S_5\right\}$ adaptively.
At scale $i$, this module first concatenates $V_i$ and $S_i$, and then feed it into a convolutional layer to learn an attention map $AM_i$.
Finally, aggregated multi-modal features $FM_i$ is obtained.
All operations in this module are formulated by:

\begin{equation}
AM_i = h\left(g\left(\left[V_i, S_i\right]\right)\right)
\end{equation}
\begin{equation}
FM_i = AM_i \odot V_i + \left(1-AM_i\right) \odot S_i
\end{equation}
where $\left[\cdot\right]$ denotes the concatenation operation, $g\left(\cdot\right)$ is a convolutional layer with kernel size $1\times1\times\left(C_i^V+C_i^S\right)\times C_i^S$ and $h\left(\cdot\right)$ is a non-linear activation function.
$\odot$ denotes the element-wise multiplication.
Through this module, a set of fused multi-modal features $FM=\left\{FM_2, FM_3, FM_4, FM_5\right\}$ are generated, which serve as the multimodal multi-scale features of a document.
Then, FPN~\cite{DBLP:conf/cvpr/LinDGHHB17} (feature pyramid network) is applied on $FM$ and provides enhanced representations.

\subsection{Relation Module}
\label{sec_rm}

\begin{figure}[t]
	\centering
	\includegraphics[width=\textwidth]{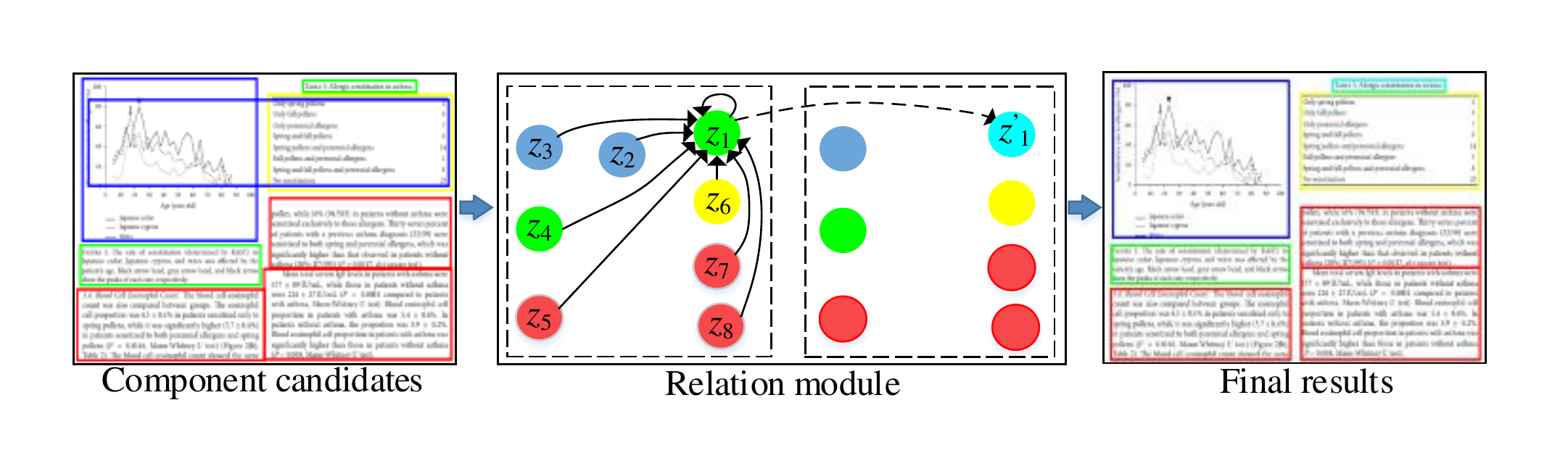}
	\caption{Illustration of relation module. It captures relations between component candidates, and thus improves detection results (remove false {\color{blue}Figure} prediction, correct {\color{cyan}Table Caption} label and adjust {\color{red}Paragraph} coordinates).
		The colors of semantic labels are: \color{blue}{Figure}, \color{red}{Paragraph}, \color{green}{Figure Caption}, \color{yellow}{Table}, \color{cyan}{Table Caption}. \color{black}{Best viewed in color.}}
	\label{figure:rm_illustration}
\end{figure}

{
Given aggregated features $FM=\left\{FM_2, FM_3, FM_4, FM_5\right\}$, a standard object detection  or segmentation model {(\eg Mask RCNN~\cite{DBLP:conf/nips/RenHGS15})}  can be used to generate component candidates in a document.}
Previous works directly take those predictions as final results. 
However, strong relations exist between layout components. {For example, bounding boxes of {Paragraphs} in the same column should be aligned; {Table} and {Table Caption} often appear together; there is no overlap between components.
{We find that such relations can be utilized to further refine predictions, as shown in Fig~\ref{figure:rm_illustration}, \ie adjusting regression coordinates for aligned bounding boxes, correcting wrong prediction labels based on co-occurrence of components and removing false predictions based on non-overlapping property. 
Next, we show how we use GNN (graph neural network) to model component relations and how to use it to refine prediction results.}

We represent a document as a graph 
$G=\left(O, E\right)$, {where $O=\left\{o_1,\cdots,o_N\right\}$ is the node set and $E$ is the edge set. 
Each node {$o_j$} represents a component candidate generated by the object detection model previously, and each edge represents the relation between two component candidates.
Since remote regions in a document may also bear close dependencies (\eg a paragraph spans two columns), all regions constitute a neighbor relationship.
Thus, the document graph is a fully-connected graph and $E\subseteq O\times O$.
The key idea {of our relation module} is to update the hidden representations of each node by attending over its neighbors ($z_1, z_2, \cdots, z_8$$\rightarrow$$z_1'$, as shown in Fig~\ref{figure:rm_illustration}). With updated node features, we could predict its {refined} label and position coordinates.

Initially, each node, denoted by $o_j=\left(b_j, f_j\right)$, includes two pieces of information: position coordinates $b_j$ and deep features $f_j=RoIAlign(FM, b_j)$. 
In order to incorporate both of them into node representation, we construct new node feature $z_j$ as follows,
\begin{equation}
\label{q_z_j}
z_j = LayerNorm(f_j + e^{pos}_j\left(b_j\right))
\end{equation}
where $e^{pos}_j\left(b_j\right)$ is the position embedding vectors of $j$-th node.

Then, instead of explicitly specifying the relations between nodes, inspired by \cite{DBLP:conf/iclr/VelickovicCCRLB18}, we apply \textit{self-attention} mechanism to automatically learn the relations, which has already shown great success in NLP and document processing~\cite{DBLP:conf/kdd/XuL0HW020,DBLP:conf/icpr/YuLQG020,DBLP:conf/naacl/LiuGZZ19,DBLP:conf/mm/ZhangXCP0QNW20}.
Specifically,} we adopt the popular scaled dot-product attention~\cite{DBLP:conf//VaswaniSPUJGKP17} to obtain sufficient expressive power.
{Scaled dot-product attention consists of queries $Q$ and keys $K$ of dimension $d_k$, and values $V$ of dimension $d_v$.}
The output $\widehat{O}$ is obtained by weighted sum over all values in $V$, where the attention weights are obtained using $Q$ and $K$, as shown in Eq.(\ref{self_att}). Please refer to \cite{DBLP:conf//VaswaniSPUJGKP17} for details.
\begin{equation}
\label{self_att}
\widehat{O}=Attention(Q,K,V)=softmax(\frac{QK^\mathsf{T}}{\sqrt{d_{k}}})V
\end{equation}

In our context, node feature set $Z=\left\{z_1,\cdots,z_N\right\}$  serves as $K$, $Q$ and $V$ and updated node feature set $Z'=\left\{z_1',\cdots,z_N'\right\}$ is the output $\widehat{O}$.
We apply multi-head attention to further improve representation capacity of node features.

Finally, given updated node features $Z'$, refined detection results of $j$-th node ($j$-th layout component candidate) $\widetilde{o_j}=\left( \widetilde{p_j^c}, \widetilde{b_j}\right)$ is computed as,
\begin{equation}
\widetilde{p_j^c} = Softmax(Linear_{cls}\left(z'_j\right))
\end{equation}
\begin{equation}
\widetilde{b_j} = Linear_{reg}\left(z'_j\right)
\end{equation}
where $\widetilde{p_j^c}$ is the probability of belonging to $c$-th class, $\widetilde{b_j}$ is its refined regression coordinates. $Linear_{cls}$ and $Linear_{reg}$ are projection layers.

{Relation module can be easily applied to NLP-based methods.
In this case, node feature $z_j$ in Eq.(\ref{q_z_j}) is the representation of $j$-th low-level elements (\eg tokens).
Then, GNN models pairwise relations between tokens and predicts their semantic labels ($\widetilde{p_j^c}$).}

\subsection{Optimization}
\label{sec_o}
Since multimodal layout analysis can be modeled as sequence labeling or object detection tasks, their optimization losses are different.

\textbf{Layout analysis as sequence labeling.}
The loss function is formulated as, 
\begin{equation}
\label{losses_ce}
\mathcal{L}=-\frac{1}{T}\sum_{j=1}^{T}log\ \widetilde{p_j}\left(y_j\right)
\end{equation}
where, $T$ is the number of low-level elements and $y_j$ is the groundtruth semantic label of $j$-th element.

\textbf{Layout analysis as object detection.}
The loss function is generated from two parts,
\begin{equation}
\label{losses}
\mathcal{L}=\mathcal{L}_{DET} + \lambda\mathcal{L}_{RM}
\end{equation}
where $\mathcal{L}_{DET}$ and $\mathcal{L}_{RM}$ are the losses used in candidate generation process and relation module.
Both $\mathcal{L}_{DET}$ and $\mathcal{L}_{RM}$ consist of a cross entropy loss (classification) and a smooth L$_1$ loss (coordinate regression), as defined in~\cite{DBLP:conf/nips/RenHGS15}.
Hyper-parameters $\lambda$ controls the trade-off between two losses. 

\section{Experiments}

\subsection{Datasets}
\label{dataset}
All three benchmarks provide document images and their original PDFs. {Therefore, text could be directly obtained {by parsing PDFs,}} allowing the explorations of multi-modal techniques.
To compare with existing solutions on each benchmark, we {use the same evaluation metrics as used by each benchmark}. 

\textbf{Article Regions}~\cite{DBLP:conf/emnlp/SotoY19} consists of 822 document samples and 9 region classes {are annotated} ({Title, Authors, Abstract, Body, Figure, Figure Caption, Table, Table Caption and References}). The {annotation is in object detection format and the }evaluation metric is mean average precision (mAP).

\textbf{PubLayNet}~\cite{DBLP:conf/icdar/ZhongTJ19} is a large-scale document dataset recently released by IBM. It consists of 360K document samples and 5 {region classes} are annotated ({Text, Title, List, Figure, and Table}). 
{The annotation is also in object detection format.}
{They use the same evaluation metric as used in the COCO competition, \ie the mean average precision (AP) @ intersection over union (IOU) [0.50:0.95].}

\textbf{DocBank}~\cite{DBLP:conf/coling/LiXCHWLZ20} is proposed by Microsoft. It contains 500K document samples with 12 {region classes} ({Abstract, Author, Caption, Equation, Figure, Footer, List, Paragraph, Reference, Section, Table} and {Title}). {It provides token-level annotations, and use F1 score as {official} evaluation metric.}
{Also, it provides object detection annotations, supporting object detection method.}

\subsection{Implementation Details}
\label{impl_detail}
{
Document image is directly used as input for visual stream. For semantic stream, we extract embedding maps (\textit{SentGrid} and \textit{CharGrid}) from text as input, where \textit{SentGrid} is generated by pretrained BERT model~\cite{DBLP:conf/naacl/DevlinCLT19}
and \textit{CharGrid} is obtained from a word embedding layer. They all have the same channel dimension size ($C^S_0=64$).
ResNeXt-101~\cite{DBLP:conf/cvpr/XieGDTH17} is used as backbone to extract both visual and semantic features (unless otherwise specified), which are later fused by a multi-scale adaptive aggregation and feature pyramid network.

For CV-based multimodal layout analysis methods, fused features are fed into RPN, followed by RCNN, to generate component candidates. 
In RPN, 7 anchor ratios (0.02, 0.05, 0.1, 0.2, 0.5, 1.0, 2.0) are adopted to handle document elements that vary in sizes and scales.}
In relation module, dimension of each candidate is set to $1024$ and $2$ layers of multi-head attention with $16$ heads are used to model relations.
{We set $\lambda$ in Eq.(\ref{losses}) to be 1 in all our experiments.}
For NLP-based multimodal layout analysis methods, low-level elements parsed from PDFs (\eg tokens) serve as component candidates, and relation module predicts their semantic labels.

Our model is implemented under the PyTorch framework.
It is trained by the SGD optimizer with batchsize=$2$, momentum=$0.9$ and weight-decay=$10^{-4}$.
The initial learning rate is set to $10^{-3}$, which is divided by 10 every $10$ epochs on Article Regions dataset and $3$ epochs on the other two benchmarks.
The training of model on Article Regions lasts for $30$ epochs while on the other two benchmarks lasts for $6$.
All the experiments are carried out on Tesla-V100 GPUs.
Source code will be released in the near future.

\begin{table}[t]
	\centering
	\caption{Performance comparisons on Article Regions dataset}
	\label{table:AR_dataset}
	\begin{threeparttable}
		\resizebox{1\textwidth}{!}{
			\begin{tabular}{|c|ccccccccc|c|}
				\hline
				Method             & Title & Author & Abstract & Body & Figure & \begin{tabular}[c]{@{}c@{}}Figure\\ Caption\end{tabular} & Table & \begin{tabular}[c]{@{}c@{}}Table\\ Caption\end{tabular} & Reference & mAP \\ \hline
				Faster RCNN~\cite{DBLP:conf/emnlp/SotoY19} & - & 1.22& - &87.49 & - & - & - & - &-& 46.38 \\
				Faster RCNN \textit{w/ context}~\cite{DBLP:conf/emnlp/SotoY19} & -  & 10.34 & - & 93.58 & - & - & - & 30.8 & - & 70.3 \\  \hline
				Faster RCNN \textit{reimplement} & 100.0 & 51.1 & 94.8 &98.9 &94.2 &91.8 &97.3 &67.1 & 90.8 & 87.3 \\
				\begin{tabular}[c]{@{}c@{}}Faster RCNN \textit{w/ context}\\ \textit{reimplement}\end{tabular}~\cite{DBLP:conf/emnlp/SotoY19} & 100.0 & 60.5 & 90.8 &98.5 &\textbf{96.2} &91.5 &\textbf{97.5} &64.2 & 91.2 & 87.8 \\ \hline
				VSR  & \textbf{100.0} & \textbf{94} &\textbf{95}  &\textbf{99.1} &95.3 &\textbf{94.5}  &96.1  &\textbf{84.6}  &\textbf{92.3}  &\textbf{94.5}  \\ \hline
		\end{tabular}}
		\begin{tablenotes}
			\scriptsize
			\item {Note: missing entries are because those results are not reported in their original papers.}
		\end{tablenotes}
	\end{threeparttable}
\end{table}

\begin{table}[t]
	\centering
	\caption{Performance comparisons on PubLayNet dataset.}
	\label{table:publaynet}
	\resizebox{0.6\textwidth}{!}{
		\begin{tabular}{|c|c|ccccc|c|}
			\hline
			Method      & Dataset               & Text & Title & List & Table & Figure & AP \\ \hline
			Faster RCNN~\cite{DBLP:conf/icdar/ZhongTJ19} & \multirow{3}{*}{val}  &  91 &  82.6  & 88.3 & 95.4  & 93.7   & 90.2   \\
			Mask RCNN~\cite{DBLP:conf/icdar/ZhongTJ19}   &                       &  91.6 & 84  & 88.6 & 96   & 94.9  & 91   \\
			VSR       &                       &  \textbf{96.7} & \textbf{93.1} & \textbf{94.7} & \textbf{97.4} & \textbf{96.4}  & \textbf{95.7} \\ \hline
			Faster RCNN~\cite{DBLP:conf/icdar/ZhongTJ19} & \multirow{7}{*}{test} & 91.3 & 81.2  & 88.5 & 94.3  & 94.5 & 90   \\
			Mask RCNN~\cite{DBLP:conf/icdar/ZhongTJ19}   &                       & 91.7 & 82.8  & 88.7 & 94.7  & 95.5 & 90.7   \\
			DocInsightAI  & & 94.51& 88.31& 94.84& 95.77&97.52 & 94.19   \\
			SCUT & & 94.3& 89.72& 94.25 &96.62 &97.68 &94.51   \\
			SRK  &&94.65&89.98&\textbf{95.14}&\textbf{97.16}&\textbf{97.95}&94.98   \\
			SiliconMinds &&96.2&89.75&94.6&96.98&97.6&95.03   \\
			VSR       & &\textbf{96.69} & \textbf{92.27} &94.55 & 97.03 & 97.90 & \textbf{95.69}  \\  \hline
	\end{tabular}}
\end{table}

\subsection{Results}
\label{main_result}
\subsubsection{Article Regions.}
{We compare the performance of VSR on this dataset with two models: Faster RCNN and Faster RCNN  \textit{with context}~\cite{DBLP:conf/emnlp/SotoY19}. 
Faster RCNN \textit{with context} adds {limited} context (page numbers, region-of-interest position and size) as input in addition to document images.

In Table~\ref{table:AR_dataset}, we first show mAP as reported in their original papers~\cite{DBLP:conf/emnlp/SotoY19}. For fair comparison, we reimplement those two models using the same backbone (ResNet-101) and neck configuration as used in VSR.
We also report their performance after reimplementation.
We can see that our reimplemented models have much higher mAP than their original models.
We believe this is mainly because we use multiple anchor ratios {in RPN}, thus achieve better detection results on document elements with various sizes.
VSR makes full use of vision, semantics and relations between components, showing highest mAP on most classes.
On Figure and Table categories, VSR achieves comparable results and the slight performance drop will be further discussed in Sec~\ref{e_r_m}.}

\begin{table}[t]
	\centering
	\caption{Performance comparisons on DocBank dataset in F1 Score.}
	\label{table:docbank_F1}
	\resizebox{1\textwidth}{!}{
		\begin{tabular}{|c|cccccccccccc|c|}
			\hline
			Method             & Abstract & Author & Caption & Equation & Figure & Footer & List  & Paragraph & Reference & Section & Table & Title & \begin{tabular}[c]{@{}c@{}}Macro\\ Average\end{tabular} \\ \hline
			BERT$_{base}$           & 92.94    & 84.84  & 86.29   & 81.52    & 100.0  & 78.05  & 71.33 & 96.19     & 93.10     & 90.81   & 82.96 & 94.42 & 87.70                                                   \\
			RoBERTa$_{base}$        & 92.88    & 86.18  & 89.44   & 82.48    & 100.0  & 80.14  & 73.53 & 96.46     & 93.41     & 93.37   & 83.89 & 95.11 & 88.91                                                   \\
			LayoutLM$_{base}$       & 98.16    & 85.95  & 95.97   & 89.47    & 100.0  & 89.57  & 89.48 & 97.88     & 93.38     & 95.98   & 86.33 & 95.79 & 93.16                                                   \\
			BERT$_{large}$          & 92.86    & 85.77  & 86.50   & 81.77    & 100.0  & 78.14  & 69.60 & 96.19     & 92.84     & 90.65   & 83.20 & 94.30 & 87.65                                                   \\
			RoBERTa$_{large}$       & 94.79    & 87.24  & 90.81   & 83.70    & 100.0  & 83.92  & 74.51 & 96.65     & 93.34     & 94.07   & 84.94 & 94.61 & 89.88                                                   \\
			LayoutLM$_{large}$      & 97.84    & 87.83  & 95.56   & 89.74    & \textbf{100.0}  & 91.46  & 90.04 & 97.90     & 93.32     & 95.96   & 86.79 & 95.52 & 93.50                                                   \\ \hline
			X101               & 97.17    & 82.27  & 94.35   & 89.38    & 88.12  & 90.29  & 90.51 & 96.82     & 87.98     & 94.12   & 83.53 & 91.58 & 90.51                                                   \\
			X101+LayoutLM$_{base}$  & 98.15    & 89.07  & \textbf{96.69}   & 94.30    & 99.90  & 92.92  & 93.00 & 98.43     & 94.37     & 96.64   & 88.18 & 95.75 & 94.78                                                   \\
			X101+LayoutLM$_{large}$ & 98.02    & 89.64  & 96.66   & 94.40    & 99.94  & 93.52  & 92.93 & 98.44     & 94.30     & 96.70   & 88.75 & 95.31 & 94.88                                                   \\ \hline
			VSR                & \textbf{98.29}    & \textbf{91.19}  & 96.32   & \textbf{95.84}    & 99.96  & \textbf{95.11}  & \textbf{94.66} & \textbf{98.66}     & \textbf{95.05}     & \textbf{97.11}   & \textbf{89.24} & \textbf{95.63} & \textbf{95.59}                                                   \\ \hline
	\end{tabular}}
\end{table}

\begin{table}[t]
	\centering
	\caption{Performance comparisons on DocBank dataset in mAP.}
	\label{table:docbank_map}
	\resizebox{1\textwidth}{!}{
		\begin{tabular}{|c|cccccccccccc|c|}
			\hline
			Models & Abstract & Author & Caption & Equation & Figure & Footer & List & Paragraph & Reference & Section & Table & Title & mAP \\ \hline
			\begin{tabular}[c]{@{}c@{}}Faster\\ RCNN\end{tabular} &96.2&88.9&93.9&\textbf{78.1}&85.4&\textbf{93.4}&86.1&67.8&89.9&76.7&77.2&\textbf{95.3}& 86.3    \\ \hline
			VSR    &\textbf{96.3}&\textbf{89.2}&\textbf{94.6}&77.3&\textbf{97.8}&93.2&\textbf{86.2}&\textbf{69.0}&\textbf{90.3}&\textbf{79.2}&\textbf{77.5}&94.9& \textbf{87.6}     \\ \hline
	\end{tabular} }
\end{table}

\subsubsection{PubLayNet.}
{In Table~\ref{table:publaynet}, we compare the performance of VSR on this dataset with two {pure image-based} methods, Faster RCNN~\cite{DBLP:conf/nips/RenHGS15} and Mask RCNN~\cite{DBLP:conf/iccv/HeGDG17}.}
While those two models present promising results (AP\textgreater90\%) on validation dataset, VSR improves the performance on all classes and increases the final AP by 4.7\%. 
{VSR shows large performance improvements on text-related classes ({Text}, {Title} and {List}) since it also utilizes document semantics in addition to document image.} 
On test dataset (also known as \textit{leaderboard of ICDAR2021 layout analysis recognition competition\footnote[4]{https://icdar2021.org/competitions/competition-on-scientific-literature-parsing/}}), VSR surpasses all participating teams and ranks first, with $4.99\%$ increase on AP compared with Mask RCNN baseline.

\subsubsection{DocBank.}
This dataset offers both token and detection annotations. {Therefore, we could treat layout analysis task either as sequence labeling task 
or as object detection task, then compare VSR with existing solutions in both cases. }

\textbf{\textit{Layout analysis as sequence labeling}}.
Using token-level annotations, we compare VSR with BERT~\cite{DBLP:conf/naacl/DevlinCLT19}, RoBERTa~\cite{DBLP:journals/corr/abs-1907-11692}, LayoutLM~\cite{DBLP:conf/kdd/XuL0HW020}, Faster RCNN with ResNeXt-101~\cite{DBLP:conf/cvpr/XieGDTH17} and ensemble models (ResNeXt-101+LayoutLM) in Table~\ref{table:docbank_F1}. 
{Even though highest F1 score of {Caption} and {Figure} are achieved by ensemble model (ResNeXt-101+LayoutLM) and LayoutLM respectively, VSR achieves comparable results with small gaps ($\leq$ 0.37\%).} {More importantly, VSR gets the highest scores on all other classes. }
This indicates that VSR is significantly better than BERT, RoBERTa and LayoutLM architectures on document layout analysis task.

\textbf{\textit{Layout analysis as object detection}}. 
{Since both VSR and Faster RCNN with ResNeXt-101 can provide object detection results, we further compare them in object detection format using mAP as evaluation metric. 
Results in Table~\ref{table:docbank_map} show that VSR outperforms Faster RCNN on most classes, {except {Equation}, {Footer} and {Title}.}
{Overall}, VSR shows $1.3\%$ gains in final mAP.}

\subsection{Ablation Studies}
\label{ablation_exp}
{
VSR introduces multi-granularity semantics, two-stream network with adaptive aggregation, and relation module. Now we explore how each of them contributes to VSR's performance improvement on Article Regions dataset.}

\subsubsection{Effects of multi-granularity semantic features.}
{To understand whether multi-granularity semantic features indeed improve VSR's performance, we compare 4 versions of VSR (\textit{vision-only}, \textit{vision+character}, \textit{vision+sentence}, 
\textit{vision+\\character+sentence}) in Table~\ref{table:xxgrid}.
Here \textit{character} and \textit{sentence} refer to semantic features at two different granularities.} 
We can see that, introducing document semantics at each granularity alone can boost analysis performance while {combining both of them} leads to highest mAP.
This is consistent with {how humans comprehend documents. Humans can better recognize regions which require little context from characters/words (\eg {Author}) and those which need context from sentences (\eg {Table caption}).}

\begin{table}[t]
	\centering
	\caption{Effects of semantic features at different granularities.}
	\label{table:xxgrid}
	\resizebox{1\textwidth}{!}{
	\begin{tabular}{|ccc|ccccccccc|c|}
		\hline
		 \multicolumn{1}{|c|}{\multirow{2}{*}{Vision}} & \multicolumn{2}{c|}{Semantics}       & \multirow{2}{*}{Title} & \multirow{2}{*}{Author} & \multirow{2}{*}{Abstract} & \multirow{2}{*}{Body} & \multirow{2}{*}{Figure} & \multirow{2}{*}{\begin{tabular}[c]{@{}c@{}}Figure\\ Caption\end{tabular}} & \multirow{2}{*}{Table} & \multirow{2}{*}{\begin{tabular}[c]{@{}c@{}}Table\\ Caption\end{tabular}} & \multirow{2}{*}{Reference} & \multirow{2}{*}{mAP} \\ \cline{2-3}
		 \multicolumn{1}{|c|}{} & \multicolumn{1}{c|}{Char} & Sentence & & &  &  & & & & & &  \\ \hline
		$\surd$ & & & 100.0 & 51.1 & 94.8 &98.9 &94.2 &91.8 &97.3 &67.1 & 90.8 & 87.3  \\
		$\surd$ & $\surd$& & 100.0&71.4 &\textbf{96.5} &98.9 &95.6 &\textbf{93.6} &96.9 &68.6 &89.9 & 90.2 \\
		$\surd$ & &$\surd$ &100.0 &60.2 &95.5 &\textbf{99.0} &\textbf{97.8} &93.2 &98.9 &\textbf{73.0} &91.2& 89.8  \\
		$\surd$ &$\surd$ &$\surd$ &\textbf{100.0} &\textbf{84.3} &96.1 &98.7 &95.7 &92.5 &\textbf{99.4} &71.4 &\textbf{92.4} & \textbf{92.3} \\ \hline
	\end{tabular}}
\end{table}

\begin{table}[t]
	\centering
	\caption{Effects of two-stream network with adaptive aggregation.}
	\label{table:two_stream_vs_single_stream}
	\resizebox{1\textwidth}{!}{
		\begin{tabular}{|c|c|ccccccccc|c|c|}
			\hline
			\multicolumn{2}{|c|}{Method}                                                                      & Title & Author & Abstract & Body & Figure & \begin{tabular}[c]{@{}c@{}}Figure\\ Caption\end{tabular} & Table & \begin{tabular}[c]{@{}c@{}}Table\\ Caption\end{tabular} & Reference & mAP & FPS \\ \hline
			\multirow{2}{*}{\begin{tabular}[c]{@{}c@{}}Single-stream\\ at input level\end{tabular}}    & R101 &       94.7&58.7&82.7&98.1&97.9&\textbf{96.3}&91.8&63.7&91.5&86.2 & 19.07   \\ \cline{2-2}
			& R152 & 100.0 & 50.5& 85.3& 97.9& \textbf{98.0}& 94.4& 93.3& 62.6& 90.5& 85.8& 18.15   \\ \hline
			\multirow{2}{*}{\begin{tabular}[c]{@{}c@{}}Single-stream\\ at decision level\end{tabular}} & R101 &       99.5&67.6&95.1&{98.8}&95.0&93.2&96.6&70.7&91.3&89.8 & \textbf{19.79}  \\ \cline{2-2}
			& R152 & 100.0& 80.2& 91.0& \textbf{99.4}& 96.0& 92.4& 98.3&\textbf{73.8}& 91.7& 91.4& 16.43  \\ \hline
			{VSR} & R101 & \textbf{100.0} &\textbf{84.3} &\textbf{96.1} &98.7 &95.7 &92.5 &\textbf{99.4} &{71.4} &\textbf{92.4} & \textbf{92.3} & 13.94 \\ \hline
			
	\end{tabular}}
\end{table}

\subsubsection{Effects of two-stream network with adaptive aggregation.}
We propose a two-stream network with adaptive aggregation module to combine vision and semantics of document.
To verify its effectiveness, we compare our VSR with its multimodal single-stream counterparts in Table~\ref{table:two_stream_vs_single_stream}.
{Instead of using extra stream to extract semantic features, single-stream networks directly use text embedding maps and concatenate them with visual features at input-level~\cite{DBLP:journals/corr/abs-2002-06144} or decision-level~\cite{DBLP:conf/cvpr/YangYAKKG17}.}
\cite{DBLP:journals/corr/abs-2002-06144} performs concatenation fusion in the input level and shows worse performances, while
\cite{DBLP:conf/cvpr/YangYAKKG17} fuses multimodal features in the decision level and achieves impressive performances ($89.8$ mAP).
{VSR first extracts visual and semantic features separately using two-stream network, and then fuses them adaptively.
This leads to highest mAP ($92.3$).
At the same time, VSR can run {at real-time (13.94 frames per second).}
We also experiment on larger backbone (ResNet-152) and reach consistent conclusions as shown in Table~\ref{table:two_stream_vs_single_stream}.

\begin{table}[t]
	\centering
	\caption{Effects of relation module.}
	\label{table:erm}
	\resizebox{1\textwidth}{!}{
		\begin{tabular}{|c|c|ccccccccc|c|}
			\hline
			\multicolumn{2}{|c|}{Method} & Title & Author & Abstract & Body & Figure & Figure caption & Table & Table caption & Reference & mAP \\ \hline
			\multicolumn{1}{|l|}{\multirow{2}{*}{Faster RCNN}} & w/o RM & 1 & 51.1 & 94.8 & 98.9 & \textbf{94.2} & 91.8 & 97.3 & 67.1 & 90.8 & 87.3 \\ \cline{2-2}
			\multicolumn{1}{|l|}{} & w/ RM & \textbf{1} & \textbf{88.4} & \textbf{99.1} & \textbf{99.1} & 85.4 & \textbf{92.6} & \textbf{98.0} & \textbf{79.2} & \textbf{91.6} & \textbf{92.6} \\ \hline
			\multirow{2}{*}{VSR} & w/o RM & 1 & 84.3 & \textbf{96.1} & 98.7 & \textbf{95.7} & 92.5 & \textbf{99.4} & 71.4 & \textbf{92.4} & 92.3 \\ \cline{2-2}
			& w/ RM & \textbf{1} & \textbf{94} & 95 & \textbf{99.1} & 95.3 & \textbf{94.5} & 96.1 & \textbf{84.6} & 92.3 & \textbf{94.5} \\ \hline
		\end{tabular}
	}
\end{table}

\begin{figure}[t]
	\centering
	\includegraphics[width=0.7\textwidth]{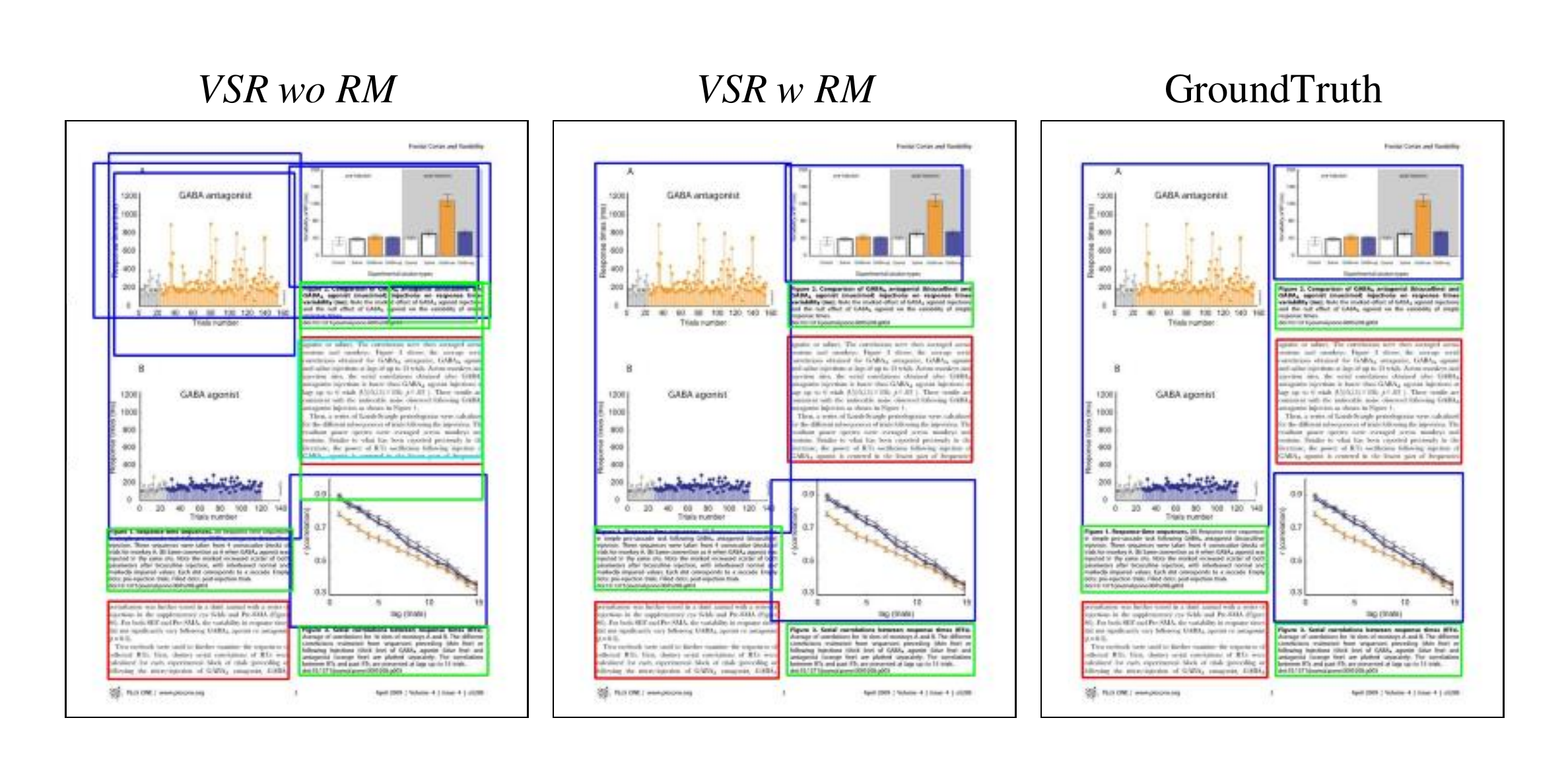}
	\caption{Qualitative comparison between \textit{VSR w/wo RM}. Introducing $RM$ effectively removes duplicate predictions and provides more accurate detection results (both labels and coordinates). The colors of semantic labels are: \color{blue}{Figure}, \color{red}{Body}, \color{green}{Figure Caption}.}
	\label{figure:result_vis}
\end{figure}

\subsubsection{Effects of relation module.}
\label{e_r_m}
To verify the effectiveness of relation module (\textit{RM}), {we compare two versions of Faster RCNN and VSR  in Table~\ref{table:erm}, \ie with \textit{RM} and without \textit{RM}.}
{Since both labels and position coordinates can be refined in \textit{RM}, both unimodal Faster RCNN and VSR show consistent improvements after incorporating relation module, with $5.3\%$ and $2.2\%$ increase respectively.
Visual examples are given in Fig~\ref{figure:result_vis}.
However, for {Figure} component, performance may slightly drop after introducting \textit{RM}.
The reason is that, while removing duplicate predictions, our relation module may also risk removing correct predictions.
But still, we see improvements on overall performances, showing the benefits of introducting relations.

\subsubsection{Limitations.}
As mentioned above, in addition to document images, VSR also requires the positions and contents of texts in the document.
Therefor, the generalization of VSR may be not good enough compared with its unimodal counterparts, which we'll address in the future.
\section{Conclusion}
\label{conclusion}
In this paper, we present a unified framework VSR for multimodal layout analysis combining vision, semantics and relations.
We first introduce semantics of document at character and sentence granularities.
Then, a two-stream convolutional network is used to extract modality-specific visual and semantic features, which are further fused in the adaptive aggregation module.
Finally, given component candidates, a relation module is adopted to model relations between them and output final results.
On three benchmarks, VSR outperforms its unimodal and multimodal single-stream counterparts significantly.
In the future, we will investigate pre-training models with VSR and extend it to other tasks, such as information extraction.

%
%
%
\bibliographystyle{splncs04}
\bibliography{mybibliography}

\end{document}